\UseRawInputEncoding

\documentclass[10pt,twocolumn,letterpaper]{article}

\usepackage{cvpr}              

\usepackage{graphicx}
\usepackage{amsmath}
\usepackage{amssymb}
\usepackage{booktabs}
\usepackage{diagbox}
\usepackage{bm}
\usepackage{xcolor}
\usepackage{algorithm}
\usepackage{algorithmic}
\usepackage{verbatim}
\usepackage{marvosym}
\usepackage[pagebackref,breaklinks,colorlinks]{hyperref}

\usepackage[capitalize]{cleveref}
\crefname{section}{Sec.}{Secs.}
\Crefname{section}{Section}{Sections}
\Crefname{table}{Table}{Tables}
\crefname{table}{Tab.}{Tabs.}

\def\cvprPaperID{\#5} 
\def\confName{CVPR IMW}
\def\confYear{2023}

\begin{document}

\title{Language Guided Local Infiltration for Interactive Image Retrieval}

\author{Fuxiang Huang, Lei Zhang$^{(}$\textsuperscript{\Letter}$^)$\\
Learning Intelligence \& Vision Essential (LiVE) Group\\
School of Microelectronics and Communication Engineering, Chongqing University, China\\
{\tt\small \{huangfuxiang, leizhang\}@cqu.edu.cn}}

\maketitle

\begin{abstract}
Interactive Image Retrieval (IIR) aims to retrieve images that are generally similar to the reference image but under the requested text modification. The existing methods usually concatenate or sum the features of image and text simply and roughly, which, however, is difficult to precisely change the local semantics of the image that the text intends to modify. To solve this problem, we propose a Language Guided Local Infiltration (LGLI) system, which fully utilizes the text information and penetrates text features into image features as much as possible. Specifically, we first propose a Language Prompt Visual Localization (LPVL) module to generate a localization mask which explicitly locates the region (semantics) intended to be modified. Then we introduce a Text Infiltration with Local Awareness (TILA) module, which is deployed in the network to precisely modify the reference image and generate image-text infiltrated representation. Extensive experiments on various benchmark databases validate that our method outperforms most state-of-the-art IIR approaches.
\end{abstract}

\section{Introduction}
Image Retrieval (IR) \cite{Survey2022} is a crucial computer vision task that serves as the foundation for a variety of applications, such as product search \cite{guo2019attentive,DML2016}, internet search \cite{sharma2019retrieving,2017Large}, person re-identification~\cite{2019Joint,liu2020hierarchical}, and face recognition~\cite{DLFR2014}. The most prevalent paradigms in image retrieval take either image or text as the input query to search for items of interest, commonly known as single-modal retrieval (i.e., image $\rightarrow$ image) \cite{PWCF2020, Brown2020eccv, Auto2020, Domain2021} and cross-modal retrieval (i.e., text $\rightarrow$ image)  \cite{CrossModalRetrieval2019, Multi2020, Hu2021, Chun2021, MultiSentence2021}. However, it is impractical for these paradigms in some real-world scenarios when users cannot precisely describe their intentions by only a single image or a single text. Therefore, one of the biggest challenges of building image retrieval systems is gaining the ability to understand the users' intentions accurately.

\begin{figure}[t]
\centering
\includegraphics[width=0.45\textwidth]{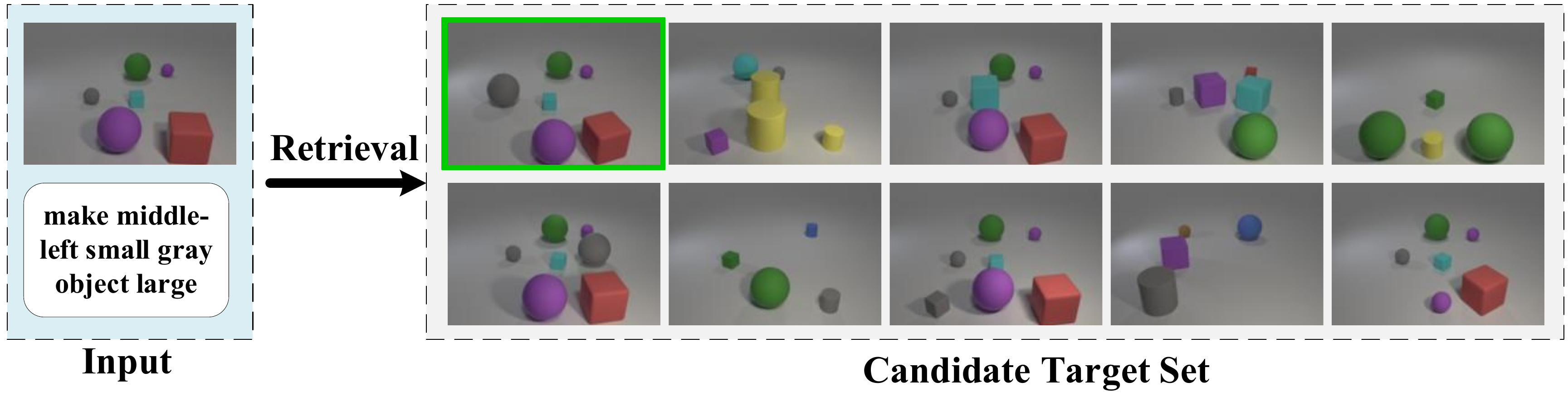}
\caption{Overview of the Interactive Image Retrieval, i.e., image+text $\rightarrow$ image. The input consists of a query image and a text describing the intended modification for searching the target image. The output is a set of images that are most similar to the query image+text. Green box means the ground-truth.}
\label{fig1}
\end{figure}

To overcome the aforementioned limitations, a challenging Interactive Image Retrieval (IIR) (i.e., Image Retrieval with Text Feedback or Language-guided Image Retrieval) \cite{vo2019composing, chen2020VAL, shin2021rtic, tautkute2021i} is proposed, which introduces text expressing a requested modification to convey users' intention together with a reference image. Fig. \ref{fig1} shows an illustration of the IIR. Currently, language-guided image retrieval methods focus on feature fusion and learning the similarity between the compositional feature (i.e., reference image+text) and the target image feature. The widely used feature fusion method is to simply fuse the features of the reference image and text. \textit{However, it is difficult to accurately and explicitly modify the local semantics of the query image by the query text via direct feature fusion}. For example, in Fig. \ref{fig1}, the desired modification (``make middle-left small gray object large'') is only related to local attributes or objects in the image (i.e., middle-left small gray object). Therefore, in contrast to other approaches, \textit{we argue that some related local regions in the image to the query text should be paid extra attention and other unrelated regions should be as unchanged as possible}.

To this end, we propose a Language Guided Local Infiltration (LGLI) system, which contains two novel components. 1) In order to discover the local area being modified, we propose a Language Prompt Visual Localization (LPVL) module to generate a localization mask of the reference image. As shown in Fig. \ref{fig2}, when the purpose is ``\textit{make small gray object large}'', we can first locate the position of the ``\textit{gray object}'' via the LPVL module and generate the localization mask. 2) We propose a Text Infiltration with Local Awareness (TILA) module, which is deployed in each layer of the network to precisely modify the local semantic of the reference image and then generate image-text infiltrated representations. TILA module learns the infiltrated representation of the \textit{masked} image and modification text without maliciously tampering with unrelated semantics. The main contributions of this work are three-fold:
\begin{figure*}[t]
\centering
\includegraphics[width=0.9\textwidth]{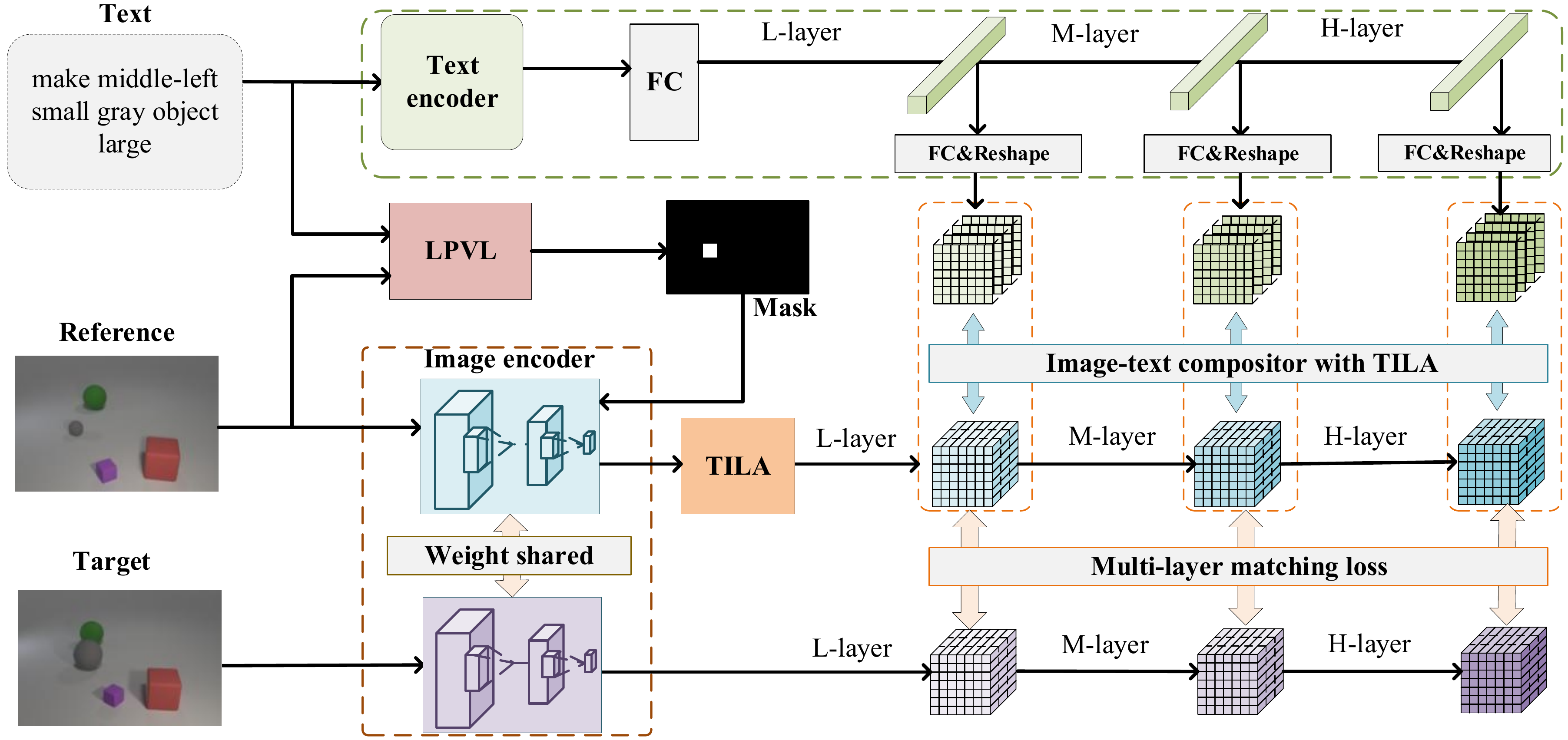}
\caption{Overview of the Language Guided Local Infiltration (LGLI) system, which contains four components: (1) an image encoder to learn visual representation, (2) a text encoder to learn text representation, (3) an LPVL module to generate a localization mask of the reference image, and (4) an image-text compositor, i.e. TILA, to generate the image-text compositional representation (infiltrated feature). Best viewed in color.}
\label{fig2}
\end{figure*}

\begin{itemize}
\item  We propose a Language Guided Local Infiltration (LGLI) system for interactive image retrieval tasks. This allows us to accurately change what the modification text intends to modify in the reference image but preserve other contents. Fig. \ref{fig2} depicts the overall architecture of our method. 
\item We propose a Language Prompt Visual Localization (LPVL) module to generate localization masks of reference images and a Text Infiltration with Local Awareness (TILA) module to learn the infiltrated representation of the reference image and modification text. These two modules can achieve the modification intent as accurately as possible.
\item The proposed method is extensively evaluated on benchmark datasets and consistently outperforms most state-of-the-art IIR approaches.
\end{itemize}

\section{Related Work}
\textbf{Interactive Image Retrieval (IIR)} aims to incorporate user text into an image retrieval system to modify or refine the image retrieval results based on the users' expectations. Recently, to solve the IIR problem, Vo \textit{et al.} \cite{vo2019composing} first propose a TIRG network by feature composition of image and text features. Dodds \textit{et al.} \cite{MAAF2020} propose a Modality-Agnostic Attention Fusion (MAAF) model to combine image and text features. LBF \cite{ho2020CQIR} uses locally bounded features to capture the detailed relationship between the text and entity in the image. A unified Joint Visual Semantic Matching (JVSM) model \cite{chen2020JVSM} is proposed to jointly associate visual and textual modalities in a shared discriminative embedding space via compositional losses. A Visiolinguistic Attention Learning (VAL) framework \cite{chen2020VAL} is proposed to fuse image and text features via attention learning at varying representation depths. Anwaar \textit{et al.} \cite{anwaar2021compositional} argue that the query image and the target image lie in a common complex space and propose a ComposeAE model to learn the composed representation of query image and text. Lee \textit{et al.} \cite{lee2021CoSMo} propose a Content-Style Modulation (CoSMo), which introduces two modules based on deep neural networks: the content and style modulators. Gu \textit{et al.} \cite{gu2021DIM} propose a new method to improve TIRG based on the contrastive self-supervised learning method, i.e., Deep InfoMax (TIGR-DIM). Joint Prediction Module (JPM) \cite{JPM2021} is proposed to align the intrinsic relationship among the query image, the target image and the modification text. Zhang \textit{et al.} propose a Joint Attribute Manipulation and Modality Alignment Learning (JAMMAL) \cite{2020Joint} method. Shin \textit{et al.} \cite{shin2021rtic} propose a Residual Text and Image Composer (RTIC), which combines the graph convolutional network (GCN) with existing composition methods. Tautkute \textit{et al.} \cite{tautkute2021i} propose a SynthTripletGAN framework for interactive image retrieval with synthetic query expansion. Hou \textit{et al.} \cite{Hou2021ICCV} introduce a architecture for learning attribute-driven disentangled representations to solve the interactive image retrieval task. Wen \textit{et al.} \cite{CLVC-Net2021} propose a Comprehensive Linguistic-Visual Composition Network (CLVC-Net), which designs two composition modules: fine-grained local-wise composition module and fine-grained global-wise com-position module. Goenka \textit{et al.} \cite{FashionVLP2022} propose a new vision-language transformer based model, FashionVLP , that brings the prior knowledge contained in large image-text corpora to the domain of fashion image retrieval, and combines visual information from multiple levels of context to effectively capture fashion-related information. Huang \textit{et al.} \cite{GA2022} propose a plug-and-play Gradient Augmentation (GA) based regularization approach to improve the generalization of IIR models.

\textit{However, the existing approaches usually directly fuse the features of reference image and modification text, which treat different local area indiscriminately, and the fact that the text described intention generally expresses the local semantic of a query image is overlooked}. Therefore, to address the text guided local-awareness issue, in this work, we propose a Language Guided Local Infiltration system. In contrast, the proposed method can explicitly discover and modify the text related local semantic in the image, but keep other unrelated semantics as unchanged as possible. 

\section{Proposed Method}
\begin{figure*}
\centering
\includegraphics[width=0.9\textwidth]{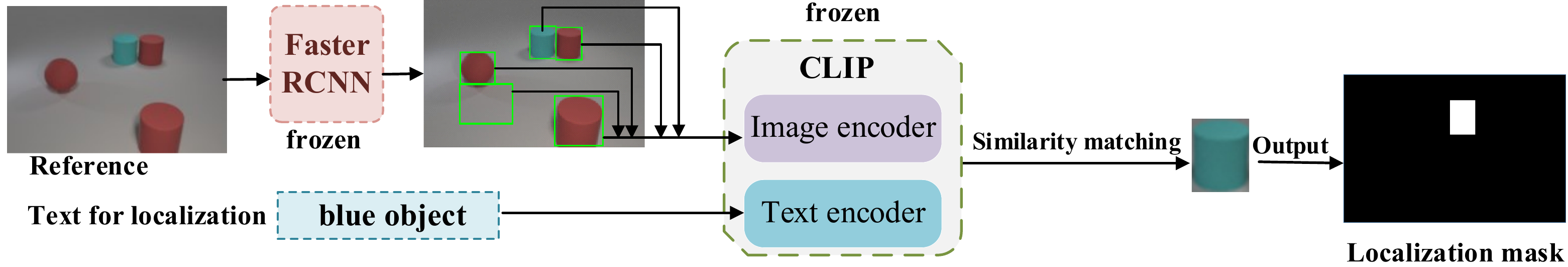}
\caption{Overview of the Language Prompt Visual Localization (LPVL). Best viewed in color.}
\label{fig3}
\end{figure*}
\subsection{Problem Definition and Model Framework}
In IIR task, we are given a reference image and a modification text that describes what content should be modified in the image. The objective of our primary task is to learn an image-text compositional embedding that encodes the information required to retrieve the target image of interest, which should reflect the changes specified by the modification text.
Let $I_r$ denote the reference image, $T$ denote the modification text, and $I_t$ denote the target image. Given an input pair $(I_r, T)$, we aim to learn image-text compositional features $f = \mathcal{C}(I_r, T; \Theta)$ that are well-aligned with the target image feature $f_t=\mathcal{F}(I_t; \Theta)$ by maximizing their similarity as,
\begin{equation}
\max_\mathbf{\Theta} \quad \kappa(\mathcal{C}(I_r, T; \Theta),\mathcal{F}(I_t; \Theta)),
\label{eq1}
\end{equation}
where $\mathcal{C}(\cdot)$ and $\mathcal{F}(\cdot)$ denote the feature composer and image feature extractor, respectively. $\kappa(\cdot,\cdot)$ denotes the similarity kernel, which is implemented as a dot product. $\Theta$ denotes the learnable parameter of the whole model.

To achieve this goal, we propose a Language Guided Local Infiltration (LGLI) system. Fig. \ref{fig2} illustrates the overall pipeline to obtain the image-text infiltrated feature. Our proposed method contains four components: 1) an image encoder to learn visual representation, 2) a text encoder to learn text representation, 3) a Language Prompt Visual Localization (LPVL) module to generate localization masks of reference images, and 4) an image-text compositor with Text Infiltration with Local Awareness (TILA) to generate the image-text compositional representation (infiltrated feature). Model parameters are optimized by multi-layer matching loss.

Without loss of generality, to encapsulate the visual contents into discriminative representations, we employ an image encoder, i.e., a standard CNN, for image representation learning. In this work, we extract the feature maps from multiple convolution layers to capture the visual information of different granularities. Concretely, the image features are obtained from three different levels inside the CNN, i.e., low, mid, and high layers. To represent the semantics of modification texts, we first utilize the LSTM as a text encoder to extract text representations. Then we construct MLPs to represent the text features of low, mid, and high layers.

After obtaining the feature representations of texts and images, our primary goal is to learn the image-text compositional presentation. In contrast to previous approaches, we argue that the local region in the image wanted to modify should be paid extra attention, and other regions should be as unchanged as possible. Therefore, we design a LPVL module detailed in Sec. \ref{LPVL} to generate localization masks of reference images and an image-text compositor via TILA module detailed in Sec. \ref{TILA} to learn the infiltrated representation of the reference image and modification text.

\subsection{Language Prompt Visual Localization (LPVL)} \label{LPVL}

In this paper, we propose to pre-locate the region to be modified in the reference image. We expect to use the location information as the input of the image-text compositor to achieve the purpose of modifying only the desired location content and preserving other information.

Specifically, we propose a Language Prompt Visual Localization (LPVL) module to generate localization masks of reference images. Fig. \ref{fig3} illustrates the pipeline of our proposed LPVL. The role of the LPVL module is to get the desired localization mask while inputting the reference image and modification text. For example, when the purpose is ``\textit{make blue object purple}'', we can locate the position of the ``\textit{blue object}'' in advance through the LPVL module and generate the localization mask. In this paper, we define ``\textit{blue object}'' as the localization text $T_L$ and all localization texts comes from the corresponding modification text $T$. In our LPVL module, we use Faster-RCNN \cite{Faster2015} model pre-trained on the COCO dataset to detect the regions of interest (i.e., object). CLIP \cite{2021Learning} model pre-trained on a dataset of 400 millions image-text pairs collected from the internet is used to extract language and visual representations. Specifically, Given a reference image $I_r$, we first apply a pre-trained Faster-RCNN to obtain $K$ regions of interest. The formulas are written as:
\begin{equation}
O = \{o^1, o^2,..., o^K\} = FasterRCNN(I_r),
\label{eqroi}
\end{equation}
where $FasterRCNN(\cdot)$ denotes the pre-trained Faster RCNN model and $O$ is the set of detected regions of interest.  Then each region of interest is represented as a $512$-dimensional feature vector through the pre-trained image encoder $CLIP_{img}$. The localization text $T_L$ is also represented as a $512$-dimensional feature vector through the pre-trained text encoder $CLIP_{txt}$. The features of the regions of interest and the modification text are represented as:
\begin{equation}
\left\{\begin{array}{rcl}
\begin{aligned}
&f_o^{i} = CLIP_{img}(o^i), \quad i=\{1, 2, ..., K\}\\
&f_{t_l} = CLIP_{txt}(T_L)).
\end{aligned}
\end{array} \right.
\label{eq2}
\end{equation}

In order to capture the exact local region to be modified, we calculate the similarity between each region and the modification text and select the most similar region to generate a localization mask, i.e., the localization object $LO$ is calculated as:
\begin{equation}
LO = \arg \max_{(o^{i}\in O)} \quad \kappa(f_o^{i},f_{t_l}).
\label{eq3}
\end{equation}
where $\kappa(\cdot,\cdot)$ present the similarity function, e.g., cosine similarity.

\begin{figure*}
\centering
\includegraphics[width=0.9\textwidth]{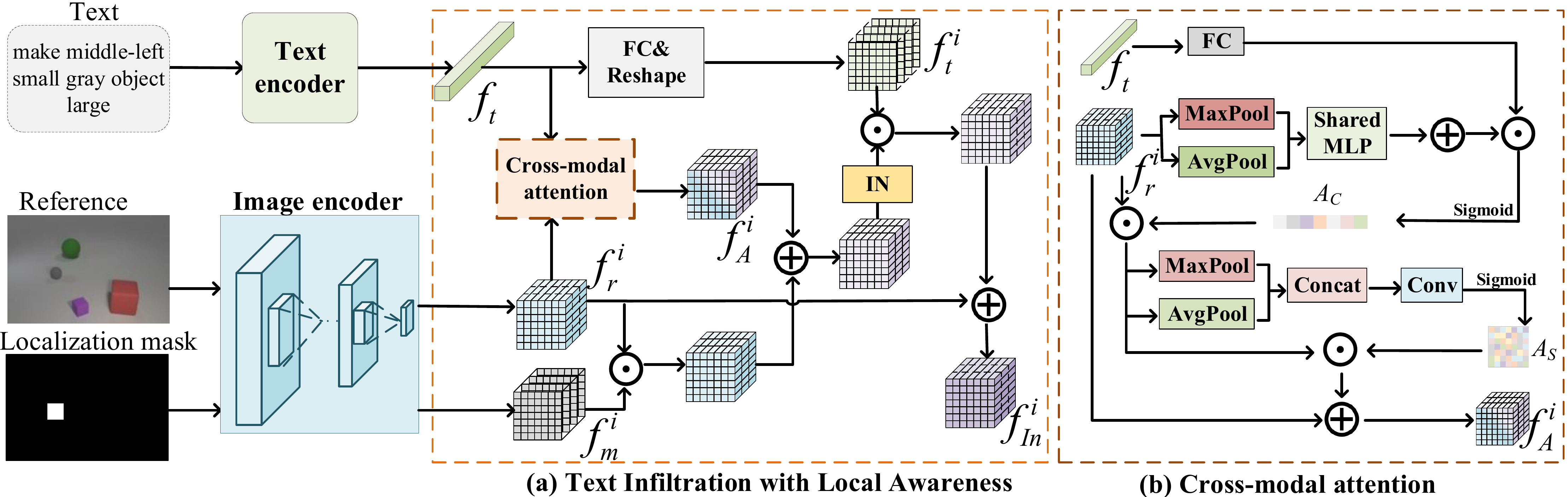}
\caption{Overview of the proposed Text Infiltration with Local Awareness (TILA) module. $\bigodot$ and $\bigoplus$ stand for the Hadamard product and element-wise addition, respectively. IN denotes instance normalization. Best viewed in color.}
\label{fig4}
\end{figure*}

\subsection{Text Infiltration with Local Awareness (TILA)} \label{TILA}
In order to accurately change the local semantics of the image to be specified and preserve other unrelated local semantics, we propose a Text Infiltration with Local Awareness (TILA) module for the image-text compositor. Fig. \ref{fig4} illustrates the overall pipeline of TILA. The key insight is to capture the local semantics and achieve specified change via the localization mask and modification text. Meanwhile, we expect that other regions cannot be inadvertently changed.

After generating the localization mask $I_m$ from the LPVL module, we leverage the modification text and localization mask to jointly change the local semantic of the reference image. Specifically, we feed the reference image into the image encoder to extract the low-, mid-, and high-level image representations, i.e., $f_r^{L}$, $f_r^{M}$, and $f_r^{H}$. Similarly, the mask representations can also be extracted via the image encoder. The text representations are extracted by the text encoder. For brevity, we use $f_r^{i}$, $f_m^{i}$, and $f_t^{i}$ to represent the image, mask, and text features, respectively, which are formulated as:
\begin{equation}
\left\{\begin{array}{rcl}
\begin{aligned}
&f_r^{i} = \mathcal{F}(I_r; \theta_{CNN}^i),\\
&f_m^{i} = \mathcal{F}(I_m; \theta_{CNN}^i), i\in{L, M, H}\\
&f_t^{i} = Reshape(Linear(E(T))),
\end{aligned}
\end{array} \right.
\label{eq4}
\end{equation}
where $E(\cdot), Linear(\cdot), Reshape(\cdot)$ are text encoder, linearization and reshape operators, respectively. $f_r^{i}, f_m^{i}, f_t^{i}\in \mathbb{R}^{c_i \times h_i \times w_i}$. $c_i$, $h_i$, and $w_i$ refer to the number of the channel, height, and width in the $i$-layer of the image encoder, respectively. To achieve text guided local-aware image feature, we first compose the features of the reference image and localization mask and obtain visual features enhanced by local information. This operation is equivalent to increasing the weight of the specified region so that the specified region is easier to modify than others. Besides, considering that the localization mask may be inaccurate, we propose a cross-modal attention mechanism to gradually adjust the modified regions. As shown in Fig. \ref{fig4} (b), the cross-modal attention contains a channel attention module and a spatial attention module, which is similar as CBAM \cite{CBAM2018} but the difference is that we consider the information of two modalities (image vs. text). In the channel attention module, we first aggregate spatial information of an image feature map by using both average-pooling and max-pooling operations, generating two different spatial context descriptors, i.e., $f_{Avg}^{Ci} = AvgPool(f_r^{i})$ and $f_{Max}^{Ci} = MaxPool(f_r^{i})$, where $f_{Avg}^{Ci}, f_{Max}^{Ci} \in \mathbb{R}^{C\times1\times1}$ . Both descriptors are then forwarded to a shared MLP network. Then the image features and text features are combined to produce our channel attention map $A_C^i$.
\begin{equation}
\begin{aligned}
A_C^i = & \sigma((MLP(f_{Avg}^{Ci}) + MLP(f_{Max}^{Ci}))) \\
      &\odot Reshape(Linear(f_t^{i}))).
      \end{aligned}
\label{eqAC}
\end{equation}
where $\sigma$ denotes the sigmoid function. Similarly, we generate a spatial attention map $A_S$ by utilizing the inter-spatial relationship of features. We aggregate channel information of a feature map by using two pooling operations, generating two 2D maps, i.e., $f_{Avg}^{Si} = AvgPool(A_C^i \odot f_r^{i})$ and $f_{Max}^{Si} = MaxPool(A_C^i \odot f_r^{i})$, where $f_{Avg}^{Si}, f_{Max}^{Si} \in \mathbb{R}^{1\times H\times W}$. These two 2D maps are then concatenated and convolved by a standard convolution layer to produce our spatial attention map $A_S^i$.
\begin{equation}
A_S^i = \sigma(Conv([f_{Avg}^{Si}, f_{Max}^{Si}]))
\label{eqAS}
\end{equation}
where $[\cdot, \cdot]$ denotes concatenation. Different from the channel attention, the spatial attention focuses on ¡®where¡¯ is an informative part, which is complementary to the channel attention. Finally, the weighted feature map by the cross-modal attention mechanism is expressed as
\begin{equation}
f_A^i = A_C^i \odot A_S^i \odot f_r^{i}
\label{eqA}
\end{equation}
Then, the enhanced visual features by localization mask and the weighted features by the cross-modal attention mechanism are processed by an Instance Normalization (IN). We then use a residual-like structure to infiltrate text features into image features as much as possible. The infiltrated features generated by TILA in layer $i$ are defined as:
\begin{equation}
\begin{aligned}
f_{In}^{i}= IN(f_r^{i}\odot \alpha f_m^{i}\!+\! f_A^{i})\odot f_t^{i}\!+\! f_r^{i},\; i\in{L, M, H},\\
\end{aligned}
\label{eq5}
\end{equation}
where $\alpha$ is a trade-off parameter used to control the effect of the localization mask. 

Note that our TILA module can be plugged and played in each layer of the CNN to precisely modify the reference image and generate image-text infiltrated representation. In this work, the image-text compositor is formulated by deploying the TILA module in the low, middle, and high layers of CNN. Finally, we get the concatenation of multi-level outputs after \textit{average-pooling} and reduce the dimension by an FC-layer.
\begin{equation}
f= Linear([avg(f_{In}^{L});avg(f_{In}^{M});avg(f_{In}^{H})]),
\label{eq6}
\end{equation}
where [$\cdot$] is concatenation operator and $avg(\cdot)$ is the average pooling operator. The final infiltrated feature $f$ used for image retrieval is a $512$-dimensional feature vector.
\begin{algorithm}
\renewcommand{\algorithmicrequire}{\textbf{Input:}}
\renewcommand{\algorithmicensure}{\textbf{Output:}}
\caption{\quad \textbf{Language Guided Local Infiltration}}
\label{Algorithm1}
\begin{algorithmic}[1]
\REQUIRE Training samples including reference image $I_r$, modification text $T$, and target image $I_t$; parameter $\alpha$.
\ENSURE  The localization mask $I_m$ and the parameters of the model $\Theta$.
\STATE Download the pre-trained Faster-RCNN and CLIP.\\
\STATE Obtain the regions of interest by Eq. (\ref{eqroi}).\\
\STATE Extract the features of regions of interest and localization text by Eq. (\ref{eq2}).\\
\STATE Calculate the localization object by Eq. (\ref{eq3}) and generate localization mask $I_m$.\\
\STATE Initialize the parameters of the model $\Theta$.\\
\STATE \textbf{repeat}:\\
 a. Extract multi-layer features  by Eq. (\ref{eq4}).\\
 b. Calculate the attention weight by Eqs. (\ref{eqAC},\ref{eqAS}) and obtain the weighted feature  of each layer by Eq. (\ref{eqA}).\\
 c. Obtain the infiltrated feature of each layer by Eq. (\ref{eq5}) and the final infiltrated feature by Eq. (\ref{eq6}).\\
 e. Update $\Theta$ with Eq. (\ref{eq7}) by SGD optimizer. \\
 \textbf{until} converge or reach maximal epoch.
\STATE Obtain the optimal model.
\end{algorithmic}
\end{algorithm}

\subsection{Objective Function}
In training phase, given an input pair ($I_r$, T), we aim to learn image-text compositional infiltrated features $f = \mathcal{C}(I_r, T; \Theta)$ that are well-aligned with the target image feature $f_t$. Following TIRG~\cite{vo2019composing}, we then adopt the multi-layer softmax cross-entropy loss as the standard loss:
\begin{equation}
\mathcal{L}_{CE}^l=  \frac{1}{B}\sum_{i=1}^{B}-\log\{ \frac{exp(\kappa(f_i^l,{f_{i}^{l}}^+ ))}{\sum_{j=1}^{B} exp(\kappa(f_i^l,f_j^l))}\},
\label{eq7}
\end{equation}
where $B$ is the batch size and $l$ is the layer number. ${f_{i}^{l}}^+ $ is the target image feature of the $i^{th}$ input pair.  The difference from TIRG is that we not only supervise the final output features but also supervise features in each layer by the softmax cross-entropy loss, such that the local discrimination is guaranteed. The details of the proposed algorithm are described in Algorithm \ref{Algorithm1}.

\section{Experiments}


\subsection{Datasets}
\textbf{CSS} \cite{vo2019composing} consists of 38K synthesized images with different colors, shapes and sizes. It contains about 19K training images and 19K testing images, respectively. Modification text for this dataset is fallen into three categories: adding new objects to the scene, removing objects from the image, and changing the attributes of the current objects in the image. Following the same evaluation protocol of TIRG \cite{vo2019composing}, we use the same training split and evaluate on the test set. Certain object shape and color combinations only appear in training but not in testing, and vice versa.

\textbf{Fashion200k} \cite{wu2020fashion} includes about 200K clothing images of 5 different fashion categories, namely: pants, skirts, dresses, tops, and jackets. Each sample is an image of a piece of a costume with accompanying attributes as the description (e.g., ``black lace scallop skater dress''). We use the same training split (around 172K images) as TIRG~\cite{vo2019composing}. During training, because there is no already matched reference image, modification text, and target image, pairwise images with attribute-like modifications are generated by comparing their product descriptions, i.e., the reference image and the target image have different attributes.

\textbf{MIT-States} \cite{mitstate2015} contains about 60K images. Each image is annotated with a noun and an adjective (such as ``sunny sky'' or ``diced fruit''). In total, the images are annotated using 245 unique nouns and 115 unique adjectives. Following the standard training and test splits provided by TIRG~\cite{vo2019composing}, we use 49 of the nouns for testing, and the rest is for training. This allows the model to learn about state/adjective (modification text) during training and has to deal with unseen objects presented in the test query. Pairs of images with the same object labels and different state labels are sampled to create the query image and target image. Therefore the system is to retrieve images that possess the same object but a new state compared with the query image.

\subsection{Compared Methods}
To validate the efficacy of our approach in the interactive image retrieval task, we compare with many representative multi-modal learning methods and interactive image retrieval methods, such as \textbf{Concatenation}, \textbf{Show and Tell} \cite{2015Show}, \textbf{Parameter Hashing} \cite{2016Image}, \textbf{MRN} \cite{2016Multimodal}, \textbf{Relationship} \cite{santoro2017simple}, \textbf{FiLM} \cite{perez2017film}, \textbf{TIRG} \cite{vo2019composing}, \textbf{LBF} \cite{ho2020CQIR}, \textbf{JVSM} \cite{chen2020JVSM}, \textbf{MAAF} \cite{MAAF2020}, \textbf{JAMMAL} \cite{2020Joint}, \textbf{JPM} \cite{JPM2021}, \textbf{ComposeAE} \cite{anwaar2021compositional}, \textbf{CoSMo} \cite{lee2021CoSMo}, \textbf{TIRG-DIM} \cite{gu2021DIM}, \textbf{CLVC-Net} \cite{CLVC-Net2021}, and \textbf{FashionVLP} \cite{FashionVLP2022}, where the last 11 of these methods are developed specifically for interactive image retrieval task over the last three years. Concatenation is a simple feature fusion method composing visual and textual features of the final layer by concatenation.

\subsection{Implementation Details}
We implement our method using PyTorch. Without loss of generality, we use the same backbone as most previous methods. Concretely, we adopt ResNet-18 as the image encoder to extract image features from and the LSTM as the text encoder to learn text representations. We use an SGD optimizer with a learning rate set to 0.01 and train the model for a maximum of 50 epochs. We empirically set $\alpha = 10^{-4}$.
The batch size $B=200$ for the CSS dataset, and $B=32$ for other datasets.
We use retrieval accuracy $R@N$ as our evaluation metric, which computes the percentage of test queries where at least one target or correctly searched image is within the top $N$ retrieved images. We also present the mean precision when $N$ is set as different values.
Besides quantitative evaluation, we also provide qualitative results.

\subsection{Results}

\textbf{Results on CSS:} Table \ref{tab1} summarizes the retrieval performance of our method and existing methods on the CSS dataset. As can be seen, our model demonstrates compelling results compared to all other alternatives. Specifically, the performance improvement of our method over the second-best method is 5.5\%.

\begin{table}[h]
\renewcommand\arraystretch{0.95}
\centering
\caption{Comparison of Performance (\%) on CSS.}
\resizebox{\linewidth}{!}{
\begin{tabular}{l|ccc|c}
\hline
Method  &R@1&R@5&R@10&Average\\
\hline
Concatenation&60.6&88.2&92.8&80.5\\
Show and Tell \cite{2015Show}&33.0&75.0&83.0&63.7 \\
MRN \cite{2016Multimodal} &60.1&--&--&--\\
Parameter Hashing \cite{2016Image}&60.5&88.1&92.9&80.5\\
Relationship \cite{santoro2017simple}&62.1&89.1&93.5&81.6\\
FiLM \cite{perez2017film}&65.6&89.7&94.6&83.3\\
JAMMAL \cite{2020Joint}&76.1&--&--&--\\
TIRG-DIM \cite{gu2021DIM}&77.0&95.6&97.6&90.1\\
TIRG \cite{vo2019composing}&78.8	&94.9	&97.3	&90.3\\
LBF \cite{ho2020CQIR} &79.2&--&--&--\\
MAAF \cite{MAAF2020}&87.8&--&--&--\\
\hline
LGLI &\textbf{93.3}&\textbf{99.0}&\textbf{99.6}&\textbf{97.3}\\
\hline
\end{tabular}}
\label{tab1}
\end{table}

\begin{figure}
\centering
\includegraphics[width=0.4\textwidth]{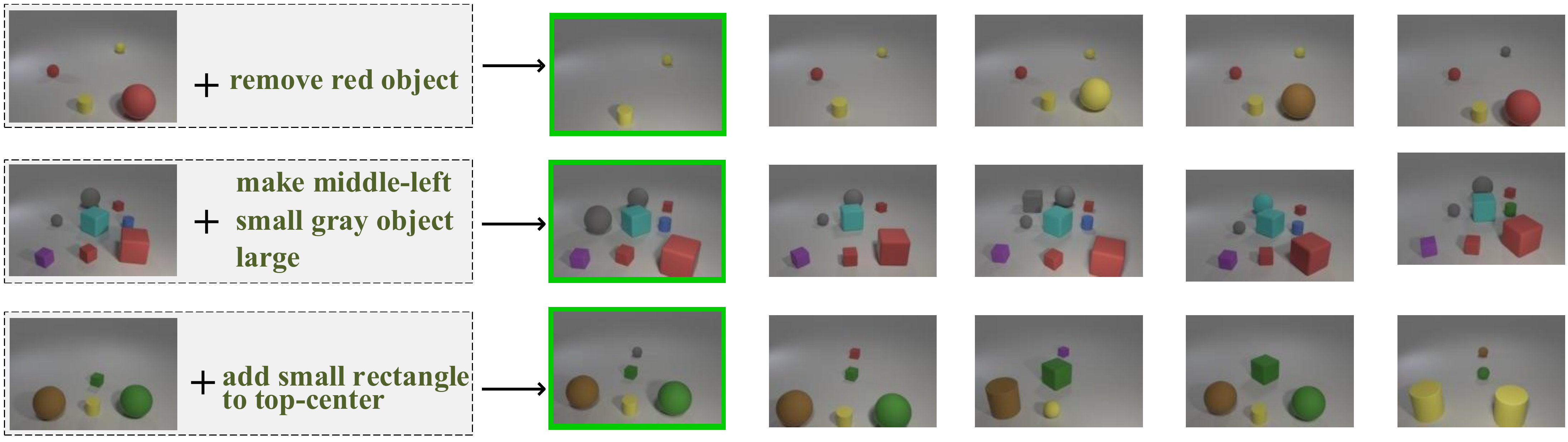}
\caption{The top 5 retrieval examples of our method on CSS dataset. The ground-truth is in green contour.}
\label{fig5}
\end{figure}

Fig. \ref{fig5} shows our qualitative results on the CSS dataset. The figure indicates our model can capture visual cues and selectively preserve and transform the reference image features according to language semantics. We observe that our model can retrieve new images that resemble the reference image, while changing certain content conditioned on text feedback, e.g., adding new objects to the scene, removing objects from the image, and changing the attributes of the current objects in the image.
\begin{table}[h]
\renewcommand\arraystretch{0.95}
\centering
\caption{Comparison of Performance (\%) on Fashion200k.}
\setlength{\tabcolsep}{1mm}{
\begin{tabular}{l|ccc|c}
\hline
Method  &R@1&R@10&R@50&Average\\
\hline
Concatenation&11.9	&39.7	&62.6	&38.1\\
Show and Tell \cite{2015Show}&12.3	&40.2	&61.8	&38.1 \\
Parameter Hashing \cite{2016Image}&12.2	&40.0	&61.7	&38.0\\
Relationship \cite{santoro2017simple}&13.0	&40.5	&62.4	&38.6\\
FiLM \cite{perez2017film}&12.9	&39.5	&61.9	&38.1\\
MRN \cite{2016Multimodal}&13.4&40.0&63.8&39.1\\
TIRG \cite{vo2019composing}&15.3	&44.3	&65.0	&41.5\\
TIRG-DIM \cite{gu2021DIM}&17.4	&43.4	&64.5	&41.8\\
JAMMAL \cite{2020Joint}&17.3	&45.3	&65.7	&42.8\\
JPM \cite{JPM2021} &19.8	&46.5	&66.6	&44.3\\
FashionVLP \cite{FashionVLP2022} &-- &49.9	&70.5   &--\\
JVSM \cite{chen2020JVSM}&19.0	&52.1	&70.0	&47.0\\
CoSMo \cite{lee2021CoSMo} &23.3	&50.4	&69.3	&47.7\\
LBF \cite{ho2020CQIR}  &17.4	&43.4	&64.5	&41.8\\
VAL	\cite{chen2020VAL}&22.9	&50.8	&72.7	&48.8\\
CLVC-Net \cite{CLVC-Net2021}&22.6	&53.0	&72.2	&49.3\\
ComposeAE \cite{anwaar2021compositional} &22.4	&55.0	&71.6	&49.7\\
\hline
LGLI &\textbf{26.5}&\textbf{58.6}&\textbf{75.6}&\textbf{53.6}\\
\hline
\end{tabular}}
\label{tab2}
\end{table}

\begin{figure}[h]
\centering
 \includegraphics[width=0.4\textwidth]{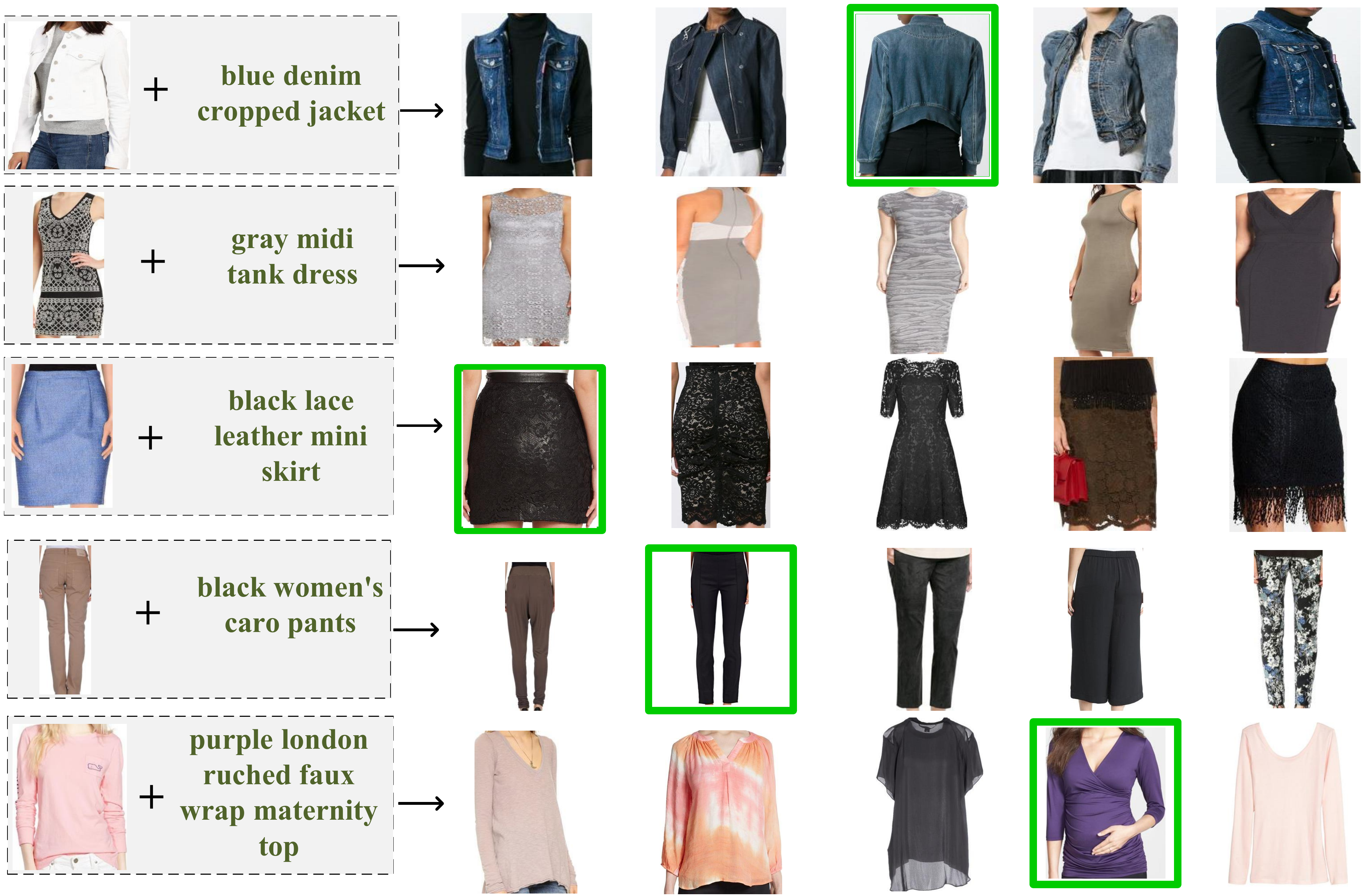}
\caption{The top 5 retrieval examples of our method on Fashion200k dataset. The ground-truth is in green contour.}
\label{fig6}
\end{figure}

\begin{table}[h]
\renewcommand\arraystretch{0.95}
\centering
\caption{Comparison of Performance (\%) on MIT-States.}
\setlength{\tabcolsep}{1.6mm}{
\begin{tabular}{l|ccc|c}
\hline
Method  &R@1&R@5&R@10&Average\\
\hline
Concatenation &11.8  &30.8  &42.1  &28.2\\
Show and Tell \cite{2015Show} &11.9  &31.0  &42.0  &28.3 \\
Parameter Hashing \cite{2016Image} &8.8  &27.3  &39.1  &25.1\\
Relationship \cite{santoro2017simple} &12.3  &31.9  &42.9  &29.0\\
FiLM \cite{perez2017film} &10.1  &27.7  &38.3  &25.4\\
MRN \cite{2016Multimodal} &11.9	&30.5	&41.0	&27.8\\
TIRG \cite{vo2019composing} &12.2  &31.9  &42.9  &29.0\\
TIRG-DIM \cite{gu2021DIM} &14.1  &33.8  &45.0  &31.0\\
JAMMAL \cite{2020Joint}&14.3	&33.2	&45.3	&30.9\\
LBF \cite{ho2020CQIR}  &14.7	&35.3	&46.6	&32.2\\
ComposeAE \cite{anwaar2021compositional} 	&13.9	&35.3	&\textbf{47.9}	&32.4\\
\hline
LGLI &\textbf{14.9}	&\textbf{36.4}	&47.7	&\textbf{33.0}\\
\hline
\end{tabular}}
\label{tabmit}
\end{table}

\begin{figure}[h]
\centering
 \centerline{\includegraphics[width=0.4\textwidth]{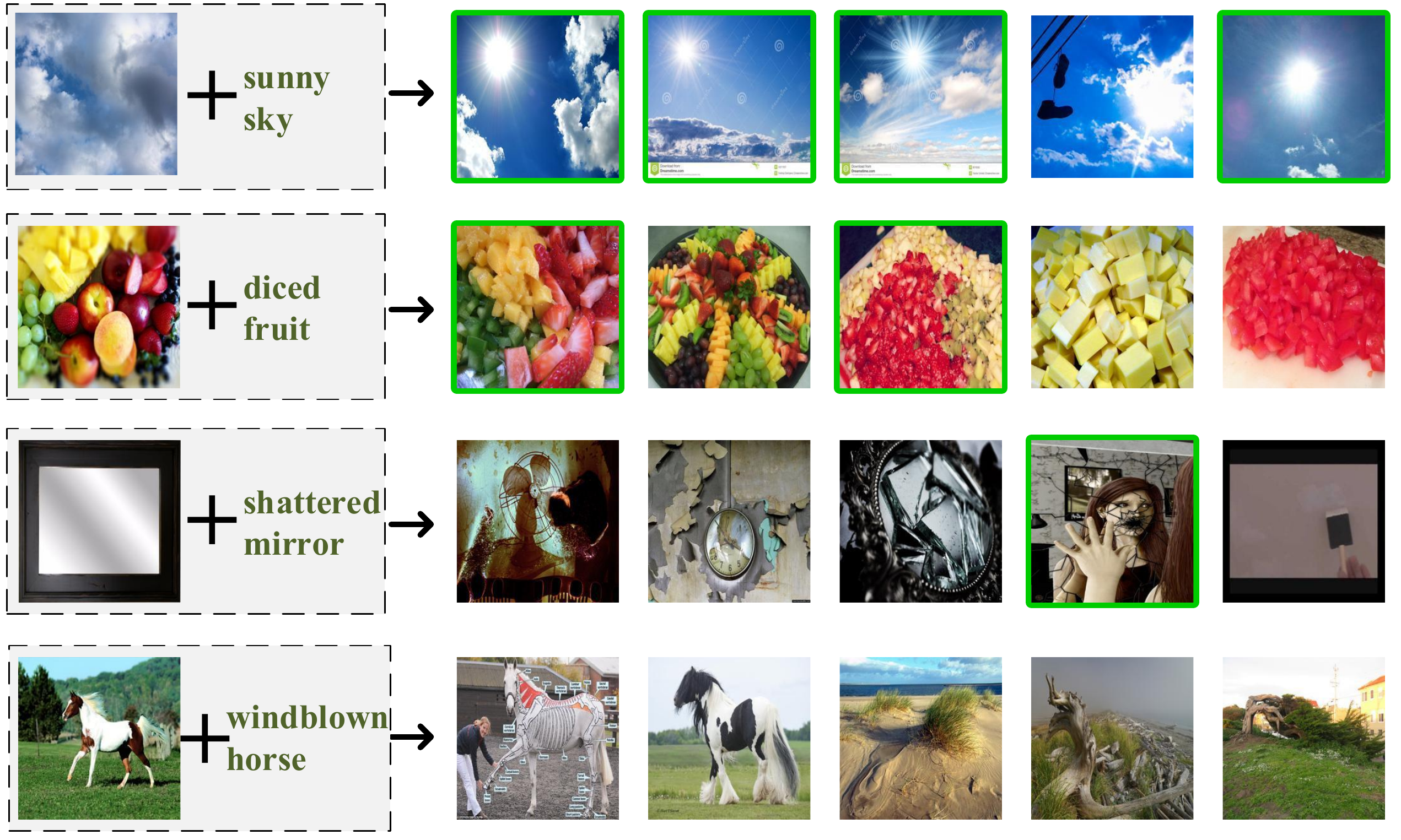}}
\caption{The top 5 retrieval examples of our method on the MIT-States dataset. The ground-truth is in green contour.}
\label{fig7}
\end{figure}

\textbf{Results on Fashion200k:} Table \ref{tab2} shows our comparison with existing methods on the Fashion200k dataset. As can be seen, our model demonstrates competitive results compared to all other alternatives. We can observe that the performance improvement of our method over the second-best method is 4.1\%, 3.6\%, 2.9\%, and 3.9\% on R@1, R@10, R@50, and average, respectively.

Fig. \ref{fig6} shows our qualitative results on the Fashion200k dataset. We notice our model can retrieve new images that resemble the reference image while changing certain content conditioned on text feedback, e.g., color, material, pattern, style, and trim.

\textbf{Results on MIT-States:} Table \ref{tabmit} shows our comparison with existing methods on the MIT-States dataset. As can be seen, our model demonstrates competitive results compared to all other alternatives. We can observe that the performance improvement of our method over the second-best method is 0.2\%, 1.1\%, and 0.6\% on R@1, R@5, and average, respectively.

Fig. \ref{fig7} shows our qualitative results on the MIT-States dataset. We notice our model can retrieve new images when the modification is easy, e.g., ``sunny sky''. However, Our model retrieval failed when the modification is abstract, e.g., ``windblown horse''.

\begin{figure*}[t]
\centering
 \centerline{\includegraphics[width=0.85\textwidth]{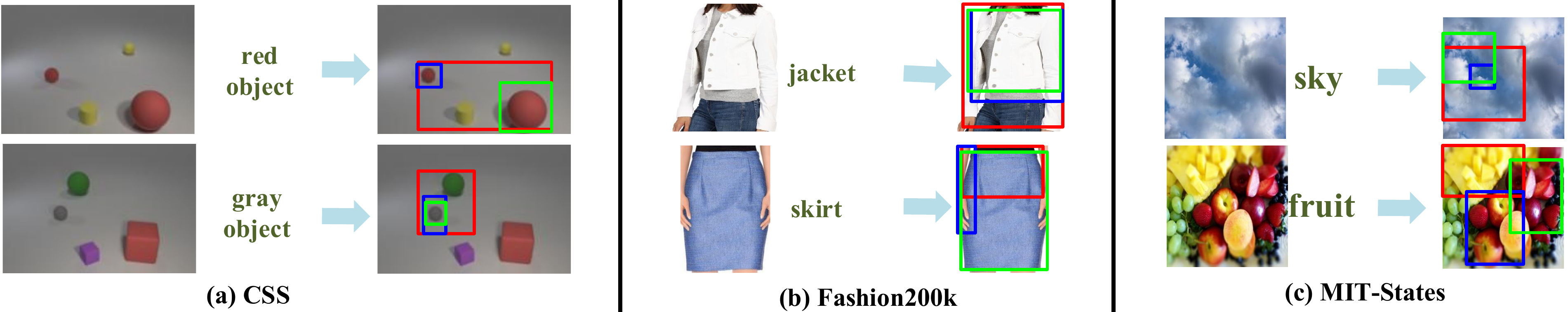}}
\caption{Localization visualization. The top 3 localization boxes are green, blue, and red, respectively. Best viewed in color.}
\label{figlocal}
\end{figure*}
Interestingly, the proposed method is usually better than others in most cases but the R@10 is lower than ComposeAE on the MIT-States dataset. This is because there are more target images in the database, i.e., the average number of target images per query on the CSS, Fashion200k and MIT-States datasets are 1, 3, and 26.7, respectively. Since there are multiple target images in the database, the correct target image can be easy to retrieved without any operation when the number of images returned by the retrieval system is sufficient.

Besides, we present the results of localization visualization through the LPVL module. As shown in Fig. \ref{figlocal}, the top 3 localization boxes of LPVL are green, blue, and red, respectively. We can see the localization boxes is usually accurate on the CSS and Fashion200k datasets, but sometimes inaccurate on the MIT-States dataset. Because MIT-States contains so many more complex categories that it is difficult to locate the region of modification. It is the reason why the improvements of our methods is not high on MIT-States.

\subsection{Ablation Study}

\begin{table}
\renewcommand\arraystretch{0.95}
\centering
\caption{Ablation Study of different components on CSS.} 
\resizebox{\linewidth}{!}{
\begin{tabular}{l|ccc|c}
\hline
Method &R@1&R@5&R@10&Average\\
\hline
LGLI &\textbf{93.3}&\textbf{99.0}&\textbf{99.6}&\textbf{97.3}\\
w/o LPVL &80.4&95.6&97.2&91.1\\
w/o CA &88.8&98.5&99.2&95.5\\
w/o TILA &74.4&93.4&96.2&88.0\\
\hline
\end{tabular}}
\label{tab4}
\end{table}

\begin{figure}
\centering
 \centerline{\includegraphics[width=0.38\textwidth]{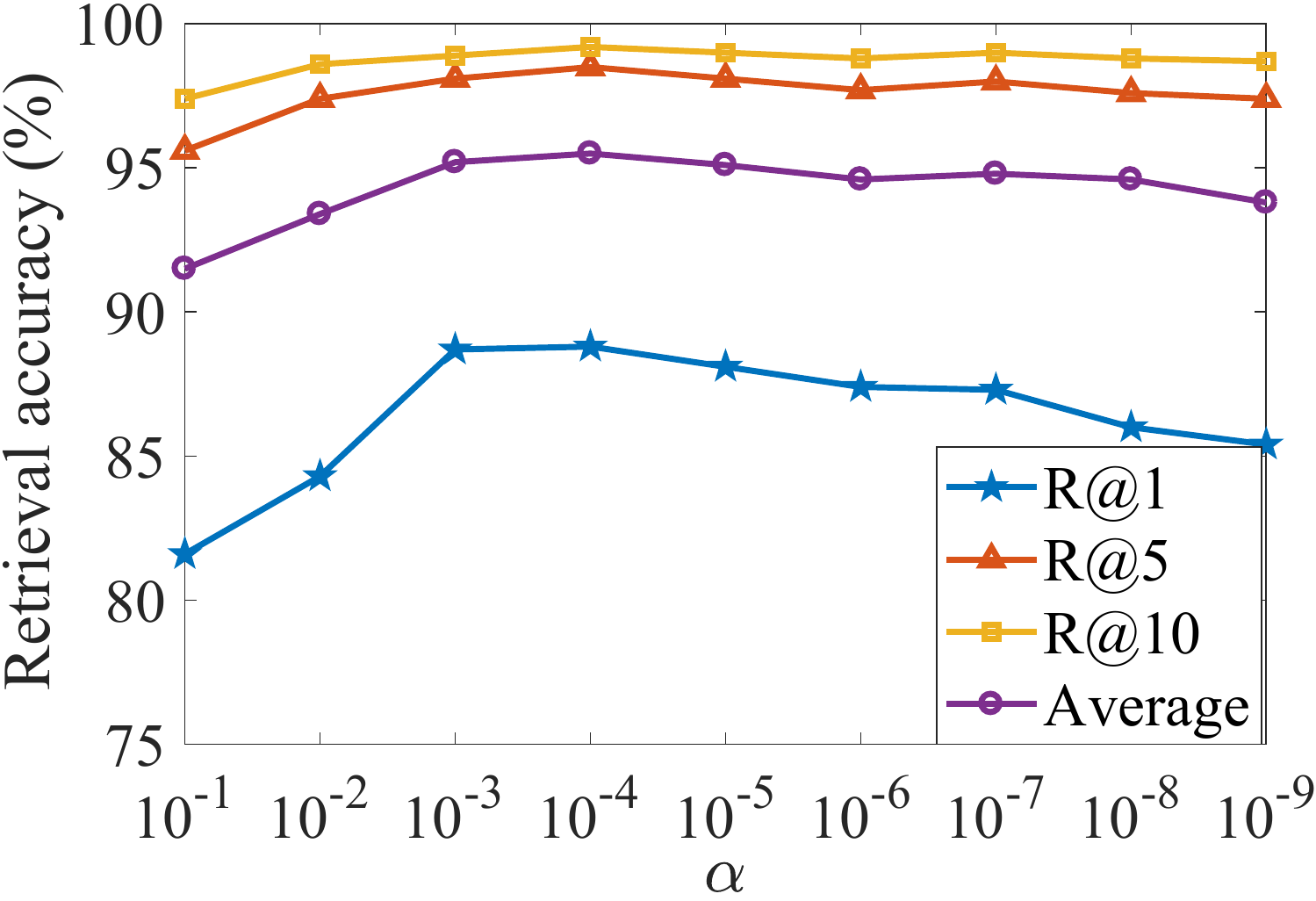}}
\caption{Sensitivity analysis of the hyper-parameter $\alpha$ on CSS.}
\label{figa}
\end{figure}
\begin{itemize}
  \item \textbf{w/o LPVL:} To investigate the effect of the LPVL module, we remove the LPVL module, i.e., not paying attention to extra location information.
  \item \textbf{w/o CA:} To explore the effect of the cross-modal attention mechanism, we remove the cross-modal attention mechanism, i.e., $f_A^i$ is replaced by $f_r^i$ in Eq. (\ref{eq5}).
  \item \textbf{w/o TILA:} To verify the role of the TILA module in the interactive image retrieval task, we replace the TILA module with concatenation to conduct a variant of our method.
\end{itemize}

Table~\ref{tab4} shows the ablation results of our LGLI, which proves the necessity of incorporating the LPVL module, cross-modal attention mechanism, and TILA module.

\subsection{Sensitivity Analysis of Hyper-parameter} We perform experiments on the CSS dataset to verify the parameter sensitivity of $\alpha$ in Eq. (\ref{eq5}). Specifically, we tune the value of ${\alpha}$ from $10^{-1}$ to $10^{-9}$. The interactive image retrieval performance variation is presented in Fig. \ref{figa}. We can observe that the performance of our proposed method is slightly sensitive to the settings of ${\alpha}$. The recommended parameter range is between $10^{-3}$ and $10^{-5}$.

\section{Conclusion}
In this paper, we propose a novel interactive image retrieval method called Language Guided Local Infiltration, which uses the location information of text attention and penetrates text features into image features as much as possible by leveraging two proposed modules, i.e., LPVL and TILA.
Specifically, in our LGLI, we first propose an LPVL module to generate a mask (i.e., localization mask) to locate the region that wanted to be modified. Then we introduce a TILA module, which can be applied to each layer of the backbone to precisely modify the reference images and generate image-text infiltrated representations. Extensive qualitative and quantitative experiments on various benchmark databases validate that our method outperforms most state-of-the-art methods.

\textbf{Acknowledgement£º}
This work was partially supported by National Natural Science Fund of China (62271090), Chongqing Natural Science Fund (cstc2021jcyj-jqX0023), National Key R\&D Program of China (2021YFB3100800), CCF Hikvision Open Fund (CCF-HIKVISION OF 20210002), CAAI-Huawei MindSpore Open Fund, and Beijing Academy of Artificial Intelligence (BAAI).
{\small
\bibliographystyle{ieee_fullname}
\bibliography{bibfile}

\begin{thebibliography}{10}\itemsep=-1pt

\bibitem{anwaar2021compositional}
Muhammad~Umer Anwaar, Egor Labintcev, and Martin Kleinsteuber.
\newblock Compositional learning of image-text query for image retrieval.
\newblock In {\em WACV}, 2021.

\bibitem{Brown2020eccv}
Andrew Brown, Weidi Xie, Vicky Kalogeiton, and Andrew Zisserman.
\newblock Smooth-ap: Smoothing the path towards large-scale image retrieval.
\newblock In {\em ECCV}, 2020.

\bibitem{chen2020JVSM}
Yanbei Chen and Loris Bazzani.
\newblock Learning joint visual semantic matching embeddings for
  language-guided retrieval.
\newblock In {\em ECCV}, 2020.

\bibitem{chen2020VAL}
Yanbei Chen, Shaogang Gong, and Loris Bazzani.
\newblock Image search with text feedback by visiolinguistic attention
  learning.
\newblock In {\em CVPR}, pages 2998--3008, 2020.

\bibitem{Chun2021}
Sanghyuk Chun, Seong~Joon Oh, Rafael~Sampaio de Rezende, Yannis Kalantidis, and
  Diane Larlus.
\newblock Probabilistic embeddings for cross-modal retrieval.
\newblock In {\em CVPR}, pages 8415--8424, 2021.

\bibitem{MAAF2020}
E. Dodds, J. Culpepper, S. Herdade, Y. Zhang, and K. Boakye.
\newblock Modality-agnostic attention fusion for visual search with text
  feedback, 2020.

\bibitem{Survey2022}
Shiv~Ram Dubey.
\newblock A decade survey of content based image retrieval using deep learning.
\newblock {\em IEEE Transactions on Circuits and Systems for Video Technology},
  32(5):2687--2704, 2022.

\bibitem{FashionVLP2022}
S. Goenka, Z. Zheng, A. Jaiswal, R. Chada, Y. Wu, V. Hedau, and P. Natarajan.
\newblock Fashionvlp: Vision language transformer for fashion retrieval with
  feedback.
\newblock In {\em CVPR}, 2022.

\bibitem{gu2021DIM}
Chunbin Gu, Jiajun Bu, Xixi Zhou, Chengwei Yao, Dongfang Ma, Zhi Yu, and Xifeng
  Yan.
\newblock Cross-modal image retrieval with deep mutual information
  maximization, 2021.

\bibitem{guo2019attentive}
Yangyang Guo, Zhiyong Cheng, Liqiang Nie, Yinglong Wang, Jun Ma, and Mohan
  Kankanhalli.
\newblock Attentive long short-term preference modeling for personalized
  product search.
\newblock In {\em ACM TOIS}, 2019.

\bibitem{ho2020CQIR}
Mehrdad Hosseinzadeh and Yang Wang.
\newblock Composed query image retrieval using locally bounded features.
\newblock In {\em CVPR}, pages 3593--3602, 2020.

\bibitem{Hou2021ICCV}
Yuxin Hou, Eleonora Vig, Michael Donoser, and Loris Bazzani.
\newblock Learning attribute-driven disentangled representations for
  interactive fashion retrieval.
\newblock In {\em ICCV}, pages 12147--12157, 2021.

\bibitem{Hu2021}
Peng Hu, Xi Peng, Hongyuan Zhu, Liangli Zhen, and Jie Lin.
\newblock Learning cross-modal retrieval with noisy labels.
\newblock In {\em CVPR}, pages 5403--5413, 2021.

\bibitem{Domain2021}
Fuxiang Huang, Lei Zhang, and Xinbo Gao.
\newblock Domain adaptation preconceived hashing for unconstrained visual
  retrieval.
\newblock {\em IEEE Transactions on Neural Networks and Learning Systems},
  2021.

\bibitem{PWCF2020}
Fuxiang Huang, Lei Zhang, Yang Yang, and Xichuan Zhou.
\newblock Probability weighted compact feature for domain adaptive retrieval.
\newblock In {\em CVPR}, pages 9579--9588, 2020.

\bibitem{GA2022}
Fuxiang Huang, Lei Zhang, Yuhang Zhou, and Xinbo Gao.
\newblock Adversarial and isotropic gradient augmentation for image retrieval
  with text feedback.
\newblock {\em IEEE Transactions on Multimedia}, 2022.

\bibitem{mitstate2015}
P. Isola, J.~J. Lim, and E.~H. Adelson.
\newblock Discovering states and transformations in image collections.
\newblock In {\em CVPR}, 2015.

\bibitem{2016Multimodal}
Jin~Hwa Kim, Sang~Woo Lee, Dong~Hyun Kwak, Min~Oh Heo, Jeonghee Kim, Jung~Woo
  Ha, and Byoung~Tak Zhang.
\newblock Multimodal residual learning for visual qa.
\newblock In {\em NIPS}, 2016.

\bibitem{lee2021CoSMo}
Seungmin Lee, Dongwan Kim, and Bohyung Han.
\newblock Cosmo: Content-style modulation for image retrieval with text
  feedback.
\newblock In {\em CVPR}, 2021.

\bibitem{liu2020hierarchical}
Zhipu Liu, Lei Zhang, and Yang Yang.
\newblock Hierarchical bi-directional feature perception network for person
  re-identification.
\newblock In {\em ACM MM}, pages 4289--4298, 2020.

\bibitem{2017Large}
H. Noh, A. Araujo, J. Sim, T. Weyand, and B. Han.
\newblock Large-scale image retrieval with attentive deep local features.
\newblock In {\em ICCV}, 2017.

\bibitem{2016Image}
H. Noh, P.~H. Seo, and B. Han.
\newblock Image question answering using convolutional neural network with
  dynamic parameter prediction.
\newblock In {\em CVPR}, 2016.

\bibitem{perez2017film}
Ethan Perez, Florian Strub, Harm de Vries, Vincent Dumoulin, and Aaron
  Courville.
\newblock Film: Visual reasoning with a general conditioning layer, 2017.

\bibitem{2021Learning}
A. Radford, J.~W. Kim, C. Hallacy, A. Ramesh, G. Goh, S. Agarwal, G. Sastry, A.
  Askell, P. Mishkin, and J. Clark.
\newblock Learning transferable visual models from natural language
  supervision.
\newblock In {\em ICML}, 2021.

\bibitem{Faster2015}
Shaoqing Ren, Kaiming He, Ross Girshick, and Jian Sun.
\newblock Faster r-cnn: Towards real-time object detection with region proposal
  networks.
\newblock In {\em NIPS}, pages 91--99, 2015.

\bibitem{santoro2017simple}
Adam Santoro, David Raposo, David G.~T. Barrett, Mateusz Malinowski, Razvan
  Pascanu, Peter Battaglia, and Timothy Lillicrap.
\newblock A simple neural network module for relational reasoning.
\newblock In {\em NIPS}, 2017.

\bibitem{sharma2019retrieving}
Rishab Sharma and Anirudha Vishvakarma.
\newblock Retrieving similar e-commerce images using deep learning, 2019.

\bibitem{Auto2020}
Yuming Shen, Jie Qin, Jiaxin Chen, Mengyang Yu, Li Liu, Fan Zhu, Fumin Shen,
  and Ling Shao.
\newblock Auto-encoding twin-bottleneck hashing.
\newblock In {\em CVPR}, pages 2815--2824, 2020.

\bibitem{shin2021rtic}
Minchul Shin, Yoonjae Cho, Byungsoo Ko, and Geonmo Gu.
\newblock Rtic: Residual learning for text and image composition using graph
  convolutional network, 2021.

\bibitem{DML2016}
Hyun~Oh Song, Yu Xiang, Stefanie Jegelka, and Silvio Savarese.
\newblock Deep metric learning via lifted structured feature embedding.
\newblock In {\em CVPR}, pages 4004--4012, 2016.

\bibitem{CrossModalRetrieval2019}
Shupeng Su, Zhisheng Zhong, and Chao Zhang.
\newblock Deep joint-semantics reconstructing hashing for large-scale
  unsupervised cross-modal retrieval.
\newblock In {\em ICCV}, pages 3027--3035, 2019.

\bibitem{DLFR2014}
Yi Sun, Xiaogang Wang, and Xiaoou Tang.
\newblock Deep learning face representation from predicting 10,000 classes.
\newblock In {\em CVPR}, pages 1891--1898, 2014.

\bibitem{tautkute2021i}
Ivona Tautkute and Tomasz Trzcinski.
\newblock I want this product but different : Multimodal retrieval with
  synthetic query expansion, 2021.

\bibitem{2015Show}
O. Vinyals, A. Toshev, S. Bengio, and D. Erhan.
\newblock Show and tell: A neural image caption generator.
\newblock In {\em CVPR}, 2015.

\bibitem{vo2019composing}
N. Vo, J. Lu, S. Chen, K. Murphy, and J. Hays.
\newblock Composing text and image for image retrieval - an empirical odyssey.
\newblock In {\em CVPR}, 2019.

\bibitem{Multi2020}
Xi Wei, Tianzhu Zhang, Yan Li, Yongdong Zhang, and Feng Wu.
\newblock Multi-modality cross attention network for image and sentence
  matching.
\newblock In {\em CVPR}, pages 10938--10947, 2020.

\bibitem{CLVC-Net2021}
H. Wen, X. Song, X. Yang, Y. Zhan, and L. Nie.
\newblock Comprehensive linguistic-visual composition network for image
  retrieval.
\newblock In {\em ACM SIGIR}, 2021.

\bibitem{CBAM2018}
S. Woo, J. Park, J.-Y. Lee, and I.~S. Kweon.
\newblock Cbam: Convolutional block attention module.
\newblock In {\em ECCV}, 2018.

\bibitem{wu2020fashion}
Hui Wu, Yupeng Gao, Xiaoxiao Guo, Ziad Al-Halah, Steven Rennie, Kristen
  Grauman, and Rogerio Feris.
\newblock Fashion iq: A new dataset towards retrieving images by natural
  language feedback, 2020.

\bibitem{MultiSentence2021}
Yanhua Yang, Lei Wang, De Xie, Cheng Deng, and Dacheng Tao.
\newblock Multi-sentence auxiliary adversarial networks for fine-grained
  text-to-image synthesis.
\newblock {\em IEEE Transactions on Image Processing}, 30:2798--2809, 2021.

\bibitem{JPM2021}
Y. Yang, M. Wang, W. Zhou, and H. Li.
\newblock Cross-modal joint prediction and alignment for composed query image
  retrieval.
\newblock In {\em ACM MM}, 2021.

\bibitem{2020Joint}
F. Zhang, M. Xu, Q. Mao, and C. Xu.
\newblock Joint attribute manipulation and modality alignment learning for
  composing text and image to image retrieval.
\newblock In {\em ACM MM}, 2020.

\bibitem{2019Joint}
Z. Zheng, X. Yang, Z. Yu, L. Zheng, Y. Yang, and J. Kautz.
\newblock Joint discriminative and generative learning for person
  re-identification.
\newblock In {\em CVPR}, 2019.

\end{thebibliography}
}

\end{document}